\documentclass{article}

    \usepackage[preprint]{neurips_2020}
\usepackage{graphicx}
\usepackage{comment}
\usepackage{amsmath,amssymb} 
\usepackage{color}
\usepackage{dsfont}

\usepackage{pifont}

\DeclareMathOperator*{\argmax}{argmax}

\usepackage{url}
\usepackage{mathtools}
\usepackage{booktabs}
\usepackage{nicefrac}
\usepackage{microtype}
\usepackage{epsfig}
\usepackage[tight,footnotesize]{subfigure}
\usepackage{amsmath}
\usepackage{amssymb}
\usepackage{amsthm}
\usepackage{bm}
\usepackage{multirow}
\usepackage{makecell}
\usepackage{footmisc}
\usepackage{marvosym}
\usepackage{latexsym}
\usepackage[utf8]{inputenc}
\usepackage{kotex}
\usepackage{titlesec}
\usepackage{caption}
\usepackage{algorithm,algpseudocode}

\newcommand{\etal}{\textit{et al}.}
\newcommand{\ie}{\textit{i}.\textit{e}., }
\newcommand{\eg}{\textit{e}.\textit{g}., }

\usepackage[bookmarksnumbered=true]{hyperref} 
\hypersetup{
     colorlinks = true,
     linkcolor = blue,
     anchorcolor = blue,
     citecolor = blue,
     filecolor = blue,
     urlcolor = blue
     }

\usepackage[utf8]{inputenc} 
\usepackage[T1]{fontenc}    
\usepackage{hyperref}       
\usepackage{url}            
\usepackage{booktabs}       
\usepackage{amsfonts}       
\usepackage{nicefrac}       
\usepackage{microtype}      

\title{Domain Adaptation without Source Data}

\makeatletter
\newcommand{\printfnsymbol}[1]{%
  \textsuperscript{\@fnsymbol{#1}}%
}
\makeatother

\author{%
   \textbf{Youngeun Kim} \\
   Yale University \\
   \texttt{youngeun.kim@yale.edu} \\
  \and
  \textbf{Donghyeon Cho}\\
  Chungnam National University \\
  \texttt{cdh12242@gmail.com} \\
  \and
  \textbf{Kyeongtak Han} \\
  Inha University \\
  \texttt{han00127@inha.edu} \\
  \and
  \textbf{Priyadarshini Panda} \\
  Yale University \\
  \texttt{priya.panda@yale.edu} \\
  \and
  \textbf{Sungeun Hong}\\
  Inha University \\
  \texttt{csehong@inha.ac.kr} \\
}

\begin{document}

\maketitle
\vspace*{-0.2in}
\begin{abstract}
{
Domain adaptation assumes that samples from source and target domains are freely accessible during a training phase.
However, such an assumption is rarely plausible in the real-world and possibly causes data-privacy issues, especially when the label of the source domain can be a sensitive attribute as an identifier.
To avoid accessing source data that may contain sensitive information,
we introduce Source data-Free Domain Adaptation (SFDA).
Our key idea is to leverage a pre-trained model from the source domain and progressively update the target model in a self-learning manner.
We observe that target samples with lower self-entropy measured by the pre-trained source model are more likely to be classified correctly. 
From this, we select the reliable samples with the self-entropy criterion and define these as class prototypes.
We then assign pseudo labels for every target sample based on the similarity score with class prototypes.
Furthermore, to reduce the uncertainty from the pseudo labeling process, we propose set-to-set distance-based filtering which does not require any tunable hyperparameters.
Finally, we train the target model with the filtered pseudo labels with regularization from the pre-trained source model. Surprisingly, without direct usage of labeled source samples, our SFDA  outperforms conventional domain adaptation methods on benchmark datasets.
Our code is publicly available at
\href{https://github.com/youngryan1993/SFDA-SourceFreeDA}{https://github.com/youngryan1993/SFDA-SourceFreeDA}.
}
\end{abstract}

\vspace*{-0.05in}

\section{Introduction}

Supervised learning has proven outstanding performance in computer vision~\cite{ren2015faster,he2016deep,chen2017deeplab,hong2019patch,cho2019key,feichtenhofer2019slowfast,kim2019cnn,lee2019bilinear}, however, performance degradation because of the differences observed from training and test environments is still problematic.
The natural solution for reducing this environmental discrepancy is to collect a set of labeled samples from test-like environments and use them for training.
Unfortunately, such rich supervision is not only time-consuming but also expensive due to the labeling costs in real world.
In such situations, unsupervised domain adaptation \cite{ganin2015unsupervised}, which can transfer knowledge from a labeled source domain to an unlabeled target domain during training, is an attractive alternative.
For this reason, considerable effort has been devoted to unsupervised domain adaptation and has shown promising results in a variety of tasks \cite{ganin2015unsupervised,long2018conditional,hsu2020progressive,tsai2018learning,hong2019unsupervised,choi2020hi}.


In general, unsupervised domain adaptation assumes that data distribution from labeled source data and unlabeled target data are related but different \cite{sugiyama2007mixture}, and all samples from both domains are freely available during the training process. 
However, this assumption is rarely possible in real-world and potentially can cause problems in terms of data privacy \cite{huawei2018security}.
Suppose that the label of source data contains bio-metric information, \eg face, fingerprint, iris pattern, or confidential information about specific individuals \cite{woodard2010periocular,naderi2016fusing,nagrani2018seeing,bae2015personal}.
From the security viewpoint, this kind of sensitive label can serve as an identifier for each sample or individual;
thus, improper disclosure of data with corresponding labels could adversely affect data providers and related organizations.
Indeed, privacy concerns have increased considerably in recent years due to the possibility of leakage of sensitive data and the vulnerability of centralized modeling. \cite{smith2017federated,yang2019federated}.

\begin{figure}[t]
\begin{center}
\def\arraystretch{0.5}
\begin{tabular}{@{}c@{\hskip 0.05\linewidth}c@{}c}
\includegraphics[width=0.42\linewidth]{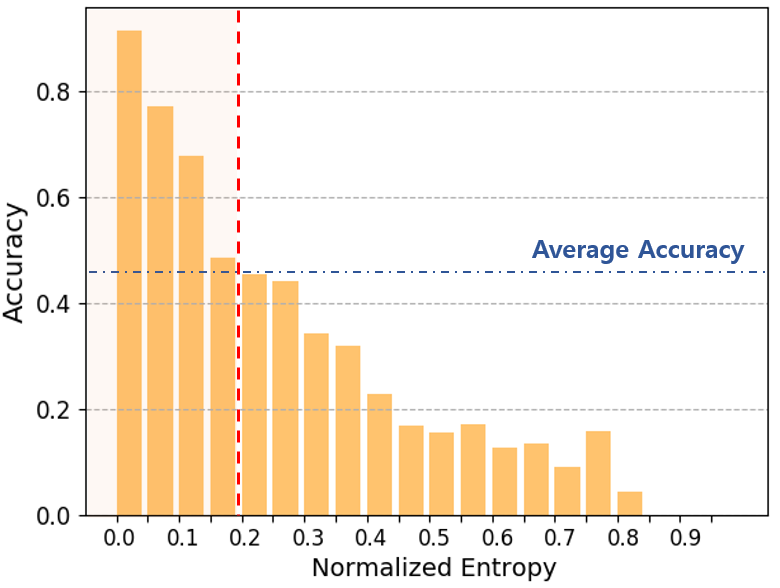} &
\includegraphics[width=0.42\linewidth]{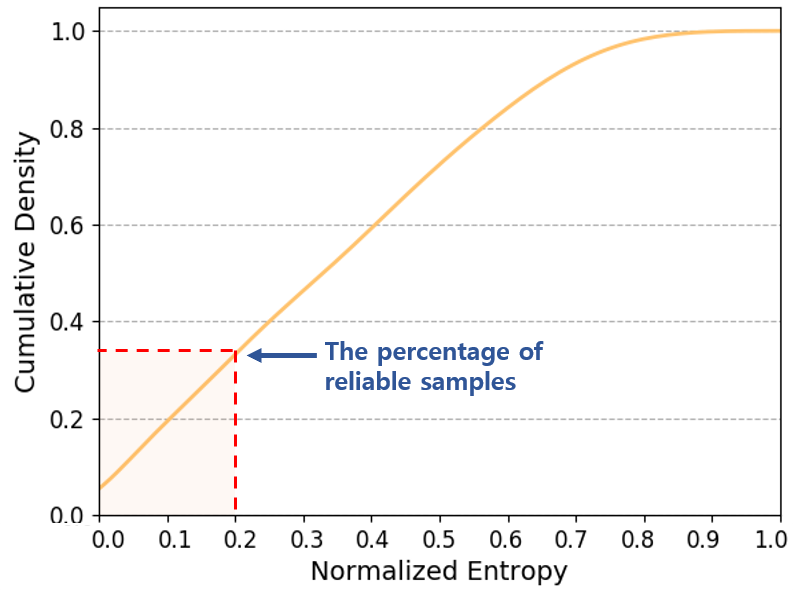} 
\\
{\hspace{2.7mm} (a) } & {\hspace{2.7mm}(b) }\\
\end{tabular}
\vspace*{-0.1in}
\end{center}
\caption{ Normalized self-entropy statistics of target samples that passed through the pre-trained source model (ResNet-50) on {Ar} $\Rightarrow$ {Cl} in Office-Home.  
(a) The smaller the self-entropy value, the higher the accuracy. 
Here, we define samples with higher accuracy than the average accuracy of the total samples as ``reliable samples'' (shaded bins).
(b) Unfortunately, these reliable samples account for a small portion of the total samples.
An in-depth analysis of the statistics regarding reliable samples will be covered in Section \ref{ssec:Analysis}.
}
\vspace*{-0.2in}
\label{fig:motivation}
\end{figure}

In this paper, we propose a novel approach that can decouple the domain adaptation process from the direct usage of  source data by leveraging pre-trained source model,  {called Source data-Free Domain Adaptation (SFDA).}
Our key idea is to update the target model using a pre-trained source model and \textit{reliable} target samples in a self-learning manner.
The natural question that arises is how to select reliable target samples from the pre-trained source model.
In domain adaptation, source and the target domains are closely related under covariate shift \cite{sugiyama2007mixture}.
Also, prediction uncertainty can be quantified by  self-entropy, \ie ${H}(x) = - \sum p(x)log(p(x))$, where smaller entropy indicates more confident prediction.
Based on this, we hypothesize that among the unlabeled target samples, samples with low self-entropy measured by the pre-trained source model are sufficiently reliable.
To verify this, we measure the self-entropy of target samples fed into a pre-trained source model and then analyze the accuracy as well as the sample distribution. 
As shown in Fig. \ref{fig:motivation},
 we treat samples with the entropy values less than $0.2$ as ``reliable samples", which accounts for about 30\% of total samples.
From the results, we can conclude that a target model can be trained with reliable target samples through the self-entropy criteria, but fatally, there are very few such samples.

\begin{figure*}[t]
     \centering
         \includegraphics[width=\textwidth]{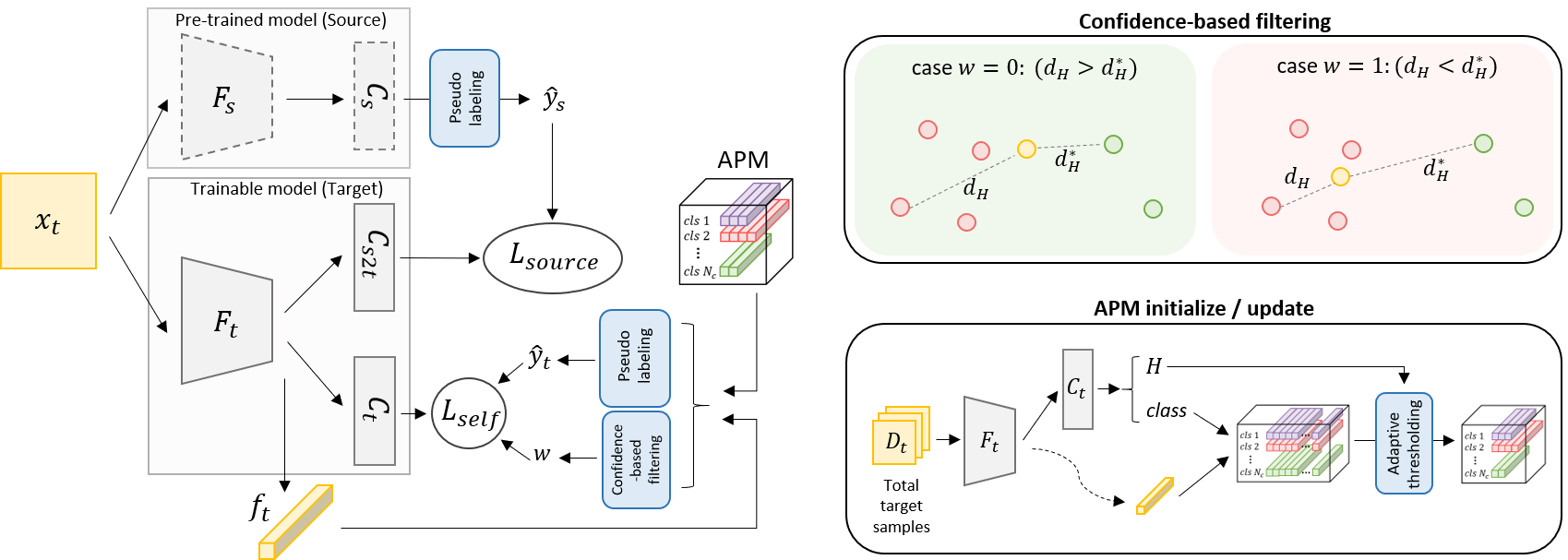}
 \caption{
 Overall flow of SFDA framework. Dashed lines indicate fixed model parameters.
 }
 \vspace*{-0.2in}
     \label{fig:Main_pipeline}
\end{figure*}

To address the issue of reliable, yet, very few target samples, we propose a new framework consisting of two parts.
One is a pre-trained model from the source domain where all the weights are frozen, and the other is a target model that is initialized from the pre-trained source model but evolves progressively by optimizing two losses (see Fig. \ref{fig:Main_pipeline}).
The first loss uses \textit{source-oriented pseudo labels} of all target samples from the pre-trained source model;
this prevents the target model from a self-biasing problem caused by the second self-learning loss.
The second loss optimizes the target model using the \textit{target-oriented pseudo labels} of target samples obtained from the trainable target network.
More precisely, we periodically store low-entropy reliable samples for each class as prototypes in a memory bank during the training process.
Then, we assign \textit{target-oriented pseudo labels} to a target sample based on the similarity between embedded features and stored class prototypes.
However, pseudo labels may not be always accurate. So, we propose a confidence-based sample filtering by measuring set-to-set distance.
As training goes on, we gradually increase the impact of the second loss, allowing our target model to adapt to the target domain in a progressive manner.

The main contributions of this work can be summarized as follows. 
(i) We tackle domain adaptation under environments with data-privacy issues. To the best of our knowledge, this is the first work on domain adaptation without any source data in training.
(ii) To decouple domain adaptation from source data, we propose a novel framework that progressively evolves based on only reliable target samples
with regularization from the source domain information. 
(iii) Although we train our target model without any source samples, our method achieves higher performance than conventional models trained with labeled source data.

\vspace{-0.1in}

\section{Source data-free domain adaptation}

\subsection{Problem setup}

The main distinction between unsupervised domain adaptation (UDA) and our proposed SFDA is that we do not use any labeled source samples during the training process.
To clarify this, we first describe the configuration of unsupervised domain adaptation and then introduce the setting of the proposed  SFDA.

\textbf{Unsupervised domain adaptation:}
The main goal of UDA is to minimize  discrepancy between a labeled source domain  $D_{s}=\{\left({x}^{i}_{s}, y^{i}_{s}\right)\}^{N_{s}}_{i=1}$ and an unlabeled target domain $D_{t}=\{{x}^{j}_{t}\}^{N_t}_{{j=1}}$, where $N_{s}$ and $N_{t}$ denote the number of source and target samples, respectively.
The main assumption of this task is that source and target samples are drawn from a different but related probability distribution $D_{s} \sim p_{s}$ and $D_{t} \sim p_{t}$, according to covariate shift \cite{sugiyama2007mixture}.

\textbf{Unsupervised domain adaptation without source data:}
Unlike UDA, the proposed SFDA protocol assumes that we cannot access samples from a source domain due to privacy issues. 
Instead of using a source dataset, this new protocol exploits a pre-trained model from the source domain.
More precisely, SFDA aims to perform unsupervised domain adaptation through the parameters of a pre-trained source model $\theta_{s}$  and the unlabeled target domain $D_{t}=\{{x}^{j}_{t}\}^{N_t}_{{j=1}}$.

\vspace{-1mm}

\subsection{SFDA framework}
\vspace{-1mm}
Figure \ref{fig:Main_pipeline} illustrates the overall flow of the proposed method.
Our SFDA framework contains two types of models: a pre-trained source model and a trainable target model. 
The source model consists of a feature extractor $F_{s}$  and a classifier $C_{s}$, and all parameters of these modules are fixed after being pre-trained from the source domain.
The trainable target model consists of a feature extractor $F_{t}$ with multi-branch classifiers, $C_{s2t}$  and $C_{t}$, and the parameters of each module are initialized by the parameter of pre-trained source model, \ie $\theta_{F_{s}}$ and  $\theta_{C_{s}}$. 
When target samples are given as input to $F_{t}$, the upper branch, \ie $C_{s2t}$, is trained with the \textit{source-oriented pseudo labels} $\hat{y_{s}}$ obtained from the pre-trained classifier $C_s$.
On the other hand, the lower branch, \ie $C_{t}$,  is trained using  \textit{target-oriented pseudo labels} $\hat{y_{t}}$ obtained by the proposed adaptive prototype memory.
Once the pseudo labels $\hat{y_{t}}$ are obtained through adaptive prototype memory, we additionally remove unreliable samples through set-to-set distance-based confidence.
Overall, we adapt the target model to the target domain using pseudo labels $\hat{y_{t}}$ in a self-learning manner, while using source knowledge $\hat{y_{s}}$ from the pre-trained source model as a regularizer.

\textbf{Adaptive prototype memory:}
Our target model updates model parameters from two types of loss functions.
The first loss function in $C_{s2t}$ aims to maintain the information of the source domain. For this, $C_{s2t}$ uses the pseudo labels obtained after inferring the target samples through the pre-trained source model, \ie  $F_{s}$  and  $C_{s}$.
Crucially, all the parameters of the pre-trained model are fixed. Hence, the loss of $C_{s2t}$ for each target sample does not change during the training process.

To evolve the target model during training, we assign pseudo labels for all target samples by using Adaptive Prototype Memory (APM).
The proposed APM consists of reliable target samples for each class called multi-prototypes.
As a first step, we compute the normalized self-entropy ${H}(x_t) = -\frac{1}{\mathrm{log}N_c} \sum {l(x_t)} \mathrm{log}({l(x_t)})$ of target samples where $l(x_t)$ denotes the predicted probability by classifier $C_t$, and $N_c$ refers to the total number of classes.
We then construct a class-wise entropy set  $\mathcal{H}_{c} = \left \{ H(x_t) | x_t \in X_c \right \}$, where $X_c$ denotes the set of samples predicted by $C_t$ as class $c$.

The next step is to choose reliable samples, called multi-prototypes, which can represent each class.
One na\"ive approach for this is to select a fixed number of samples with low self-entropy per class.
However, each class might have different entropy distribution resulting in certain classes having lower average self-entropy than others.
To address this issue, we propose an approach that can adaptively have a different number of prototypes for each class.
We first find the lowest entropy for each class, then set the largest value among them as the threshold for selecting prototypes, \ie $\eta = \max\left \{\min(\mathcal{H}_{c}) |  c \in \mathcal{C} \right \}$, where $\mathcal{C}=\left \{1, ..., N_c \right \}$ is a class set.
The main advantage of this technique is that a threshold for reliable sample selection can be obtained adaptively in the training process without any hyper-parameters.
Using this adaptive threshold, we obtain multiple prototypes for each class as follows:
\begin{equation}
    M_c = \left \{ F_t(x_t) |  x_t \in X_c, H(x_t) \leq \eta   \right \}.
    \label{eq:APM}
\end{equation}

Note that each prototype consists of an embedded feature $F_t(x_t)$.
Since multiple prototypes are later used for pseudo-labeling and confidence-based filtering, we store all prototypes into APM.
Updating APM at every training step incurs a high computational cost, so we update APM periodically.
Empirically, we update APM every 100 steps across all datasets. The entire procedure of APM is described in Algorithm 1.

\textbf{Pseudo labeling:}
Based on the multi-prototypes in APM, we can assign pseudo labels to unlabeled target samples.
Given a target sample $x_t \in \mathbb{R}^I$, we feed it into feature extractor $F_t: \mathbb{R}^I \rightarrow \mathbb{R}^E$, that yields an embedded feature $f_t \in \mathbb{R}^E$, where $I$ and $E$ stand for the dimension of input space and embedding space, respectively.
Then we compute the similarity score between the embedded feature and all prototypes in APM as: 
\begin{equation}
    s_c (x_t) = \frac{1}{|M_{c}|} \sum_{p_c \in M_c} \frac{p_c^T f_t}{\left \| p_c \right \|_2 \left \| f_t \right \|_2}.
    \label{eq:sim}
\end{equation}
where $p_c \in \mathbb{R}^E$ represents one of the multiple prototypes of the $c$ class.
Recall that each class in APM could have a different number of class prototypes.
Finally, we obtain the pseudo label by selecting the most similar class, \ie $\hat{y_t} = \argmax_{c} s_c (x_t), \forall c \in \mathcal{C}$.

\textbf{Confidence-based filtering:}
Once we obtain the pseudo labels by APM, we can train $F_{t}$ and $C_{t}$ by the conventional cross-entropy loss.
However, since we do not use any ground truth labels, there could be problems related to uncertainty or error propagation.
To get more reliable pseudo labels, we propose a filtering mechanism based on sample confidence.
Our key idea is to estimate the confidence of a pseudo label by applying set-to-set distance, which is an effective way to consider the corner cases between two sets.
The first set is a single element set consisting of one target sample, and the other set can be multiple prototypes per class.
More precisely, we obtain multiple prototypes of the most similar class $M_{t1}$  and the second similar class $M_{t2}$ for each target sample using Eq. (\ref{eq:sim}).
We then  measure the distance between the singleton set $Q = \left\{ f_t \right\}$  and $M_{t1}$ in metric space given by Hausdorff distance: 
\begin{equation}
    \begin{split}
       d_{H}(Q, M_{t1}) & = \max \left\{ 
    \adjustlimits\sup_{q\in Q} \inf_{p\in M_{t1}} {d}(q,p),   
    \adjustlimits\sup_{p\in M_{t1}} \inf_{q\in Q} {d}(q,p)
    \right\} \\
     & = \max \left\{ 
    \adjustlimits\max_{q\in Q} \min_{p\in M_{t1}} {d}(q,p),   
    \adjustlimits\max_{p\in M_{t1}} \min_{q\in Q} {d}(q,p)
    \right\} \\
     & = \max \left\{ 
    \adjustlimits \min_{p\in M_{t1}} {d}(f_t ,p),   
    \adjustlimits\max_{p\in M_{t1}}  {d}(f_t ,p)
    \right\} 
    = \max_{ p\in M_{t1}} {d}(f_t ,p).
    \end{split}
\end{equation}
Here, $d(a, b) = 1 - \frac{a \cdot b}{\left \| a \right \|_2 \left \| b \right \|_2}$ is a distance metric function.
In the second line, since $Q$, $M_{t1}$, and $M_{t2}$ are closed and discretized sets, 
we can convert the \textit{suprema} and \textit{infima} to \textit{maxima} and \textit{minima},
respectively. 
For the second most similar class, we slightly modify Hausdorff distance to select the opposite corner case as follows:
\begin{equation}
   d_{H}^* (Q, M_{t2}) = \min \left\{ 
\adjustlimits\sup_{q\in Q} \inf_{p\in M_{t2}} {d}(q,p),   
\adjustlimits\sup_{p\in M_{t2}} \inf_{q\in Q} {d}(q,p)
\right\} = \min_{p\in M_{t2}} {d}(f_t,p).
\end{equation}

{As a result, we define a reliable sample only when the prototypes of the most similar class are closer than  prototypes of the second most similar class.}
Our proposed set-to-set approach has a huge advantage, that it does not require any hyper-parameters for sample filtering.
Finally, we can assign a confidence score for each target sample as follows:
\begin{equation}
 w(x_t) =\left\{\begin{matrix} &1,  \hspace{8mm}  \textup{if} \hspace{2mm}  d_{H}(Q, M_{t1})<d_{H}^* (Q, M_{t2}) ,
\\ 
&\hspace{-29mm}0, \hspace{8mm} otherwise.
\end{matrix}\right.
\label{eq: confidence_score}
\end{equation}

\textbf{Optimization:}
Recall that, there are two trainable classifiers $C_{s2t}$  and $C_{t}$ in our target model.
The pseudo labels $\hat{y_{s}}$ obtained from the pre-trained source model are used for training $C_{s2t}$ as follows: 
\begin{equation}
   L_{source}(D_{t}) = - \mathbb{E}_{x_t \sim D_t} \sum_{c=1}^{N_c} \mathds{1}_{[c=\hat{y_s}]} log(\sigma(C_{s2t}(F_t(x_t)))),
   \label{eq:l_src}
\end{equation}
where $\sigma (\cdot)$ is a softmax function, and $\mathds{1}$ is an indicator function.
As a result, Eq. (\ref{eq:l_src}) facilitates in maintaining knowledge of the source domain and also acts as a regularizer.
On the other hand, pseudo labels $\hat{y_t}$  obtained by APM are used for training $C_{t}$  as complementary supervision:
\begin{equation}
   L_{self}(D_{t}) = - \mathbb{E}_{x_t \sim D_t} \sum_{c=1}^{N_c} w(x_t)\mathds{1}_{[c=\hat{y_t}]} log(\sigma(C_t(F_t(x_t)))).
\end{equation}

Note that we only compute the loss for confident samples using a confidence score $w(\cdot)$.
Overall, the total loss function can be formulated as follows:
\begin{equation}
   L_{total}(D_{t}) = (1-\alpha) L_{source}(D_{t}) + \alpha L_{self}(D_{t}).
  \label{eq:overall_loss}
\end{equation}
Here, $\alpha$ is the balancing parameter between the source-regularization loss $L_{source}(D_{t})$ and the self-learning loss $L_{self}(D_{t})$.
In the early stages, the pseudo label $\hat{y_t}$ is highly unstable, so we gradually increase $\alpha$ from 0 to 1. We analyze the effect of $\alpha$ in Section \ref{ssec:Analysis}.
%
Algorithm 2 describes the whole training process of our SFDA framework. In the test phase, we use the classification probability of the classifier $C_t$.
Notice that most components proposed in our work are hyper-parameter free.
The most important hyper-parameter in our work is an update period, which is analyzed in Fig. \ref{fig:WA_experiments}(b).

\begin{minipage}[t]{.45\textwidth}
    \vspace{-0.6cm}
    \begin{algorithm}[H]\small
    \caption{Initialize / Update APM}
   \textbf{Input}: unlabeled target dataset ($D_{t}=\{{x}^{j}_{t}\}^{N_t}_{{j=1}}$),
    target feature extractor ($F_t$), classifier ($C_t$)
   \\
  \textbf{Output}:  adaptive prototype memory $M$ (APM)
  \begin{algorithmic}[1]
    \State{\textbf{begin}}
    \For{$c \gets 1$ to $num\_classes$}  
        \State{$\mathcal{H}_c = $ [ ] // initialize class-wise entropy set}
    \EndFor
    \For{$t \gets 1$ to $N_t$}  
        \State{$(\hat{y_t}, H) \leftarrow (F_t,C_{t},x_t)$}
        \State {$\mathcal{H}_{\hat{y_t}}$.append[$H$]}
    \EndFor
    \State{$\eta \leftarrow (\mathcal{H}_1, ...,\mathcal{H}_{N_c})$}
    \State{$M \leftarrow [0,...,0]$}
    \For{$c \gets 1$ to $num\_classes$}  
        \State{$M_c \leftarrow (\eta, \mathcal{H}_c)$ // Eq. (\ref{eq:APM})}
    \EndFor
  \end{algorithmic}
      \label{algorithm: APM_update}
    \end{algorithm}
 \end{minipage}
 \hspace{2mm}
\begin{minipage}[t]{.52\textwidth}
    \vspace{-0.6cm}
    \begin{algorithm}[H]\small
    \caption{Optimization process}
   \textbf{Input}: fixed pre-trained source networks ($F_s$, $C_{s}$),   unlabeled target dataset ($D_{t}=\{{x}^{j}_{t}\}^{N_t}_{{j=1}}$),
   trainable target networks ($F_t$, $C_{s2t}$, $C_t$)
   \\
  \textbf{Output}:  updated  target networks $\theta = \left\{\theta_{F_t}, \theta_{C_{s2t}}, \theta_{C_{t}} \right\}$
  \begin{algorithmic}[1]
    \State{\textbf{begin}}
    \State{$(\theta_{F_t} \leftarrow \theta_{F_s})$, $(\theta_{C_{s2t}} \leftarrow \theta_{C_s})$, $(\theta_{C_{t}} \leftarrow \theta_{C_s})$ }
     \State{$M \leftarrow$ Initialize APM // Algorithm \ref{algorithm: APM_update}}
     \For{$t \gets 1$ to $max\_iter$}  
        \State {$\hat{y_s} \leftarrow (Fs,Cs)$ }
        \State {$(\hat{y_p}, w) \leftarrow (M,f_t)$ // Eq. (\ref{eq:sim}) \& Eq. (\ref{eq: confidence_score})}
        \State {$\mathcal{L}_{total} \leftarrow (x_t,\hat{y_s},\hat{y_p}, w)$ // Eq. (\ref{eq:overall_loss})}
        \State {$\theta^{(t+1)} \leftarrow \theta^{(t)} - \xi \nabla L_{total}(MiniBatch, \theta^{(t)}) $}
        \If {$t$ \% update\_period == $0$} 
         \State{$M \leftarrow$ Update $M$ // Algorithm \ref{algorithm: APM_update}}
        \EndIf
    \EndFor
    \label{algorithm: optimization}
  \end{algorithmic}
    \end{algorithm}
 \end{minipage}
 \hspace{2mm}

\section{Experiments}
\label{sec:exp}

\subsection{{Experimental setup}}
We comprehensively evaluate SFDA on three public datasets: Office-31~\cite{saenko2010adapting}, Office-Home~\cite{venkateswara2017deep}, 
and VisDA-C~\cite{peng2017visda}.
For a fair comparison with existing approaches, we follow the official UDA protocol across all the datasets.
Note,  unlike the UDA methods used for comparison in our experiments, our SFDA does not use any source samples directly during training.

\textbf{Office-31}\footnote[1]
{shorturl.at/suIY3}
consists of 4,652 images with 31 categories collected from three different domains: Amazon (A), Webcam (W), and DSLR (D).
We evaluate all methods on a total of 6 domain-transfer scenarios.
\textbf{Office-Home}\footnote[2]
{http://hemanthdv.org/OfficeHome-Dataset/}
is collected from four different domains with 65 categories and contains a total of 15,500 images: Artistic (Ar), Clipart (Cl), Product (Pr), and Real-World (Rw).
We report the performance on the 12 transfer scenarios.
\textbf{VisDA-C}\footnote[3]
{http://ai.bu.edu/visda-2017/}
is a large-scale dataset considering a synthetic-to-real scenario with 152,397 Synthetic (S) and 55,388 Real (R) images.  

\begin{table}[h!]
    \addtolength{\tabcolsep}{1.5pt}
    \centering
    \caption{Classification Accuracy (\%) on {Office-31}  ({ResNet-50})}
    \label{table:accuracy_officeic}
    \resizebox{0.75\textwidth}{!}{%
    \begin{tabular}{lccccccccc}
        \toprule
        \multirow{2}{30pt}{\centering Method}\:\:\:\:\:\:\: &  \multicolumn{7}{c}{Office-31}  \\
        \cmidrule{2-8} 
        & \:A$\Rightarrow$W\: & \:D$\Rightarrow$W\: & \:W$\Rightarrow$D\: & \:A$\Rightarrow$D\: & \:D$\Rightarrow$A\: & \:W$\Rightarrow$A\: & \:Avg\:   \\
        \midrule
        ResNet (source only)~\cite{he2016deep}   & 79.8 $\pm$ 0.6 & 98.3 $\pm$ 0.3  & 99.9 $\pm$ 0.1 & 83.8 $\pm$ 0.5 & 66.1 $\pm$ 0.3 & 65.1 $\pm$ 0.2 & 82.2  \\
        DAN~\cite{long2015learning}       & 82.6 $\pm$ 0.6 & 98.5 $\pm$ 0.1  & 100.0 $\pm$ .0 & 83.6 $\pm$ 0.5 & 67.3 $\pm$ 0.8 & 66.3 $\pm$ 0.4 & 83.1   \\
        DANN~\cite{ganin2015unsupervised} & 84.8 $\pm$ 0.9 & 98.2 $\pm$ 0.3  & 99.9 $\pm$ 0.1 & 86.3 $\pm$ 0.8 & 68.9 $\pm$ 0.4 & 66.8 $\pm$ 0.6 & 84.1   \\
        MSTN~\cite{xie2018learning}      & 86.9 $\pm$ 0.5 & 98.1 $\pm$ 0.2  & 100.0 $\pm$ .0 & 87.2 $\pm$ 0.7 & 69.6 $\pm$ 0.6 & 67.7 $\pm$ 0.6 & 84.9 \\
        MADA~\cite{pei2018multi}       & 90.0 $\pm$ 0.2 & 97.4 $\pm$ 0.1  & 99.6 $\pm$ 0.1 & 87.8 $\pm$ 0.2 & 70.3 $\pm$ 0.3 & 66.4 $\pm$ 0.3 & 85.2  \\
        CDAN~\cite{long2018conditional}       & 94.1 $\pm$ 0.1 & 98.6 $\pm$ 0.1  & 100.0 $\pm$ .0 & 92.9 $\pm$ 0.2 & 71.0 $\pm$ 0.3 & 69.3 $\pm$ 0.3 & 86.6  \\
        SAFN~\cite{xu2019larger}       & 88.8 $\pm$ 0.4 & 98.4 $\pm$ 0.0  & 99.8 $\pm$ 0.0 & 87.7 $\pm$ 1.3 & 69.8 $\pm$ 0.4 & 69.7 $\pm$ 0.2 & 85.7  \\
        \midrule
                            {SFDA w.o. CF (ours)} & 90.6 $\pm$ 0.6 & 95.3 $\pm$ 0.5  & 98.8 $\pm$ 0.3 & 92.9 $\pm$ 0.5 & 70.7 $\pm$ 0.5 & 69.8 $\pm$ 0.2 & 86.3  \\
                            {SFDA (ours)} & 91.1 $\pm$ 0.3 & 98.2 $\pm$ 0.3  & 99.5 $\pm$ 0.2 & 92.2 $\pm$ 0.2 & 71.0 $\pm$ 0.2 & 71.2 $\pm$ 0.2 & 87.2  \\
        \bottomrule
    \end{tabular}%
    }
 \vspace*{-0.1in}
\end{table}
\begin{table}[h!]
    \addtolength{\tabcolsep}{1.5pt}
    \centering
    \caption{Classification Accuracy (\%)  on {Office-Home}  ({ResNet-50})}
    \label{table:accuracy_officehome}
    \resizebox{1\textwidth}{!}{%
    \begin{tabular}{lccccccccccccc}
        \toprule
        \multirow{2}{30pt}{\centering Method}\:\:\:\:\:\:\: & \multicolumn{13}{c}{Office-Home} \\
        \cmidrule{2-14}
        & {Ar}$\Rightarrow${Cl} & {Ar}$\Rightarrow${Pr} & {Ar}$\Rightarrow${Rw} & {Cl}$\Rightarrow${Ar} & {Cl}$\Rightarrow${Pr} & {Cl}$\Rightarrow${Rw} & {Pr}$\Rightarrow${Ar} & {Pr}$\Rightarrow${Cl} & {Pr}$\Rightarrow${Rw} & {Rw}$\Rightarrow${Ar} & {Rw}$\Rightarrow${Cl} & {Rw}$\Rightarrow${Pr} & \:\:Avg\:\: \\
        \midrule
        ResNet (source only)~\cite{he2016deep}          & 42.5 & 65.8 & 74.1 & 57.4 & 63.7 & 67.7 & 55.7 & 39.1 & 72.9 & 66.1 & 46.3 & 76.9 & 60.7 \\
        DAN~\cite{long2015learning}       & 45.4 & 65.8 & 73.9 & 56.9 & 61.4 & 65.9 & 56.1 & 43.0 & 73.2 & 67.5 & 50.4 & 78.5 & 61.5 \\
        DANN~\cite{ganin2015unsupervised} & 45.7 & 66.6 & 73.4 & 58.0 & 64.9 & 68.3 & 55.9 & 42.6 & 74.1 & 66.5 & 49.7 & 77.5 & 62.0 \\
        MSTN~\cite{xie2018learning}       & 45.8 & 67.9 & 74.1 & 57.9 & 65.1 & 68.2 & 56.7 & 43.2 & 74.8 & 67.0 & 50.4 & 79.0 & 62.5 \\
        CDAN~\cite{long2018conditional}   & 50.7 & 70.6 & 76.0 & 57.6 & 70.0 & 70.0 & 57.4 & 50.9 & 77.3 & 70.9 & 56.7 & 81.6 & 65.8 \\
        SAFN~\cite{xu2019larger}   & 54.4 & 73.3 & 77.9 & 65.2 & 71.5 & 73.2 & 63.6 & 52.6 & 78.2 & 72.3 & 58.0 & 82.1 & 68.5 \\
        \midrule
        {SFDA w.o. CF (ours)} & 43.7 & 72.8 & 75.4 & 64.7 & 69.1 & 73.6 & 60.1 & 44.0 & 74.9 & 66.6 & 45.5 & 73.7 & 63.7 \\
        {SFDA (ours)}         & 48.4 & 73.4 & 76.9 & 64.3 & 69.8 & 71.7 & 62.7 & 45.3 & 76.6 & 69.8 & 50.5 & 79.0 & 65.7 \\
        \bottomrule
    \end{tabular}%
    }
 \vspace*{-0.1in}
\end{table}
\begin{table}[h!]
    \addtolength{\tabcolsep}{1.5pt}
    \centering
    \caption{Classification Accuracy (\%)  on {Visda-C}  ({ResNet-101})}
    \label{table:accuracy_visda}
    \resizebox{0.92\textwidth}{!}{%
    \begin{tabular}{lccccccccccccc}
        \toprule
        \multirow{2}{30pt}{\centering Method}\:\:\:\:\:\:\: & \multicolumn{13}{c}{Vidsa-C} \\
        \cmidrule{2-14}
        & aero & bicycle & bus & car & horse  & knife & motor & person & plant & skate & train & train & \:\:Avg\:\: \\
        \midrule
        ResNet (source only)~\cite{he2016deep}          & 88.8 & 56.1 & 67.0 & 69.3 & 92.3 & 30.3 & 87.9 & 53.1 & 81.9 & 40.0 & 82.9 & 24.8 & 64.5 \\
        {DANN~\cite{ganin2015unsupervised}} & 89.6 & 72.6 & 72.9 & 57.9 & 89.6 & 51.6 & 88.0 & 78.3 & 85.0 & 30.5 & 81.7 & 37.0 & 69.6 \\
        {MSTN~\cite{xie2018learning}}       & 89.7 & 72.6 & 75.6 & 57.4 & 91.1 & 46.5 & 88.9 & 77.4 & 85.3 & 39.2 & 81.1 & 37.8 & 70.2 \\
        MCD~\cite{saito2017maximum}       & 87.0 & 60.9 & 83.7 & 64.0 & 88.9 & 79.6 & 84.7 & 76.9 & 88.6 & 40.3 & 83.0 & 25.8 & 71.9 \\
        SFAN~\cite{xu2019larger}          & 93.6 & 61.3 & 84.1 & 70.6 & 94.1 & 79.0 & 91.8 & 79.6 & 89.9 & 55.6 & 89.0 & 24.4  & 76.1 \\
        DSBN~\cite{chang2019domain}       & 94.7 & 86.7 & 76.0 & 72.0 & 95.2 & 75.1 & 87.9 & 81.3 & 91.1 & 68.9 & 88.3 & 45.5  & 80.2 \\
        \midrule
        {{SFDA w.o. CF (ours)}}    & 89.0 & 80.5 & 77.6 & 62.3 & 92.6 & 96.1 & 87.1 & 68.3 & 79.3 & 42.8 & 68.8 & 43.0  & 73.9 \\
        {{SFDA (ours))}}            & 86.9 & 81.7 & 84.6 & 63.9 & 93.1 & 91.4 & 86.6 & 71.9 & 84.5 & 58.2 & 74.5 & 42.7  & 76.7 \\
        \bottomrule
    \end{tabular}%
    }
\end{table}

\begin{figure}[t]
\begin{center}
\def\arraystretch{0}
\begin{tabular}{@{}c@{\hskip 0.03\linewidth}c@{\hskip 0.03\linewidth}c@{\hskip 0.02\linewidth}c}
\includegraphics[width=0.32\linewidth]{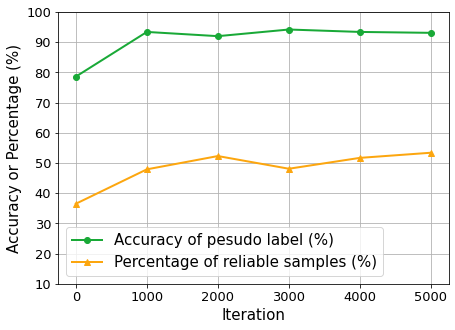} &
\includegraphics[width=0.31\linewidth]{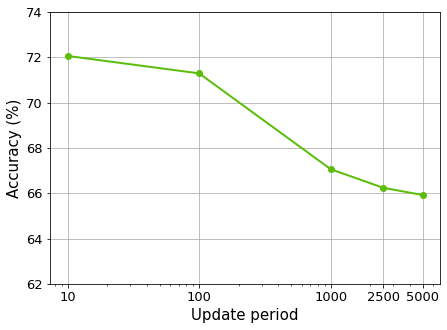}  &
\includegraphics[width=0.31\linewidth]{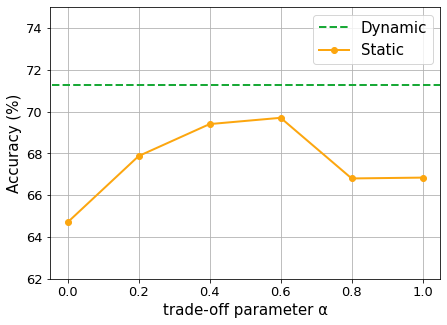}
\\
{\hspace{4.3mm} (a) } & {\hspace{4.3mm}(b) } & {\hspace{4.3mm}(c) }\\
\end{tabular}
\vspace*{-0.05in}
\end{center}
\caption
{
(a) Proportion of reliable samples (green) and accuracy of pseudo labels of reliable samples (orange) with respect to training iteration.
(b) Performance change with regard to APM update period. 
(c) Performance change with respect to a trade-off parameter in which ``static'' refers a fixed  $\alpha$ during training and ``dynamic'' indicates $\alpha$ that increases gradually during training.
All statistics are obtained from {W} $\Rightarrow$ {A} scenario in Office-31.
}
\vspace*{-0.15in}
\label{fig:WA_experiments}
\end{figure}

Following previous studies, we utilize ResNet-50 or ResNet-101 \cite{he2016deep} pre-trained on ImageNet \cite{deng2009imagenet} as a base feature extractor.
We use the same network architecture for the fixed source model and the trainable target model.
Also, we set the maximum iteration step ($max\textunderscore iter$ in Algorithm 2) to 5,000 for Office-31, 20,000 for Office-Home, and 15,000 for Visda-C. 
We use batch size 32 across all the datasets.
In our experiments, training images are resized to 256$\times$256 and 
randomly cropped to 224$\times$224 with a random horizontal flip.
We use SGD as the optimizer with a weight decay of 0.0005  and a momentum of 0.9.
The base learning rate is set to $10^{-3}$ and all fine-tuned layers in a pre-trained feature extractor are optimized with a learning rate of $10^{-4}$. Following \cite{ganin2015unsupervised}, we apply a learning rate schedule with $lr_p = lr_0(1+\alpha\cdot p)^{-\beta}$ where $lr_0$ is a base learning rate, $p$ is a relative step  that changes from 0 to 1  during training, $\alpha$ = 10, and $\beta$ = 0.75.
Note that the most important hyper-parameter of our method is an update period of APM, which we analyze in Fig. \ref{fig:WA_experiments}(b). 
Across all datasets, we update the APM module every 100 iterations.
Experiments were conducted on a TITAN Xp GPU with PyTorch implementation.

On public datasets, we compare the proposed method with the previous UDA methods including
ResNet-50 (source only)~\cite{he2016deep}, DAN~\cite{long2015learning}, DANN~\cite{ganin2015unsupervised},  MSTN~\cite{xie2018learning}, CDAN~\cite{long2018conditional}, MADA~\cite{pei2018multi}, MCD~\cite{saito2017maximum}, DSBN~\cite{chang2019domain},
and SAFN~\cite{xu2019larger}.
Importantly, early UDA methods report performance based on \textit{Caffe} while recent works are based on \textit{Pytorch} implementation.
Therefore, the performance of early works tends to be considerably lower than the recent works including ours.
For a fair comparison, we re-implement ResNet-50, DAN, DANN, and MSTN using \textit{Pytorch}. 
If the experimental setup of the previous study is the same as ours, we cite the reported results.
More implementation details can be found in supplementary material Appendix(A).

\subsection{Experimental results}

Table~\ref{table:accuracy_officeic} shows the performance comparison between our SFDA methods and previous UDA methods on Office-31. 
Following the previous protocol, we run the same setting for 5 times and report the mean and standard deviation.
In the table, ``SFDA" denotes our final model and ``SFDA w.o. CF"  refers to the SFDA variant without confidence-based filtering.
Surprisingly, even though we do not access any source samples during training, our method shows state-of-the-art accuracy.
Table~\ref{table:accuracy_officehome} shows the result on Office-Home including more severe domain transfer scenarios than Office-31.
Similar to the result on Office-31, we can see the proposed SFDA achieves a comparable performance with state-of-the-art methods.
Table~\ref{table:accuracy_visda} presents the performance of our method on Visda-C, which is the most practical and large-scale scenario.
Notably, SFDA shows a significant performance gain of 12.2\% over the source-only baseline.
Overall, our proposed SFDA consistently shows comparable performance to state-of-the-art UDA methods that use source data for training.

\begin{figure}[t]
\begin{center}
\def\arraystretch{0.5}
\begin{tabular}{@{}c@{\hskip 0.07\linewidth}c@{\hskip 0.05\linewidth}c}
\includegraphics[width=0.33\linewidth]{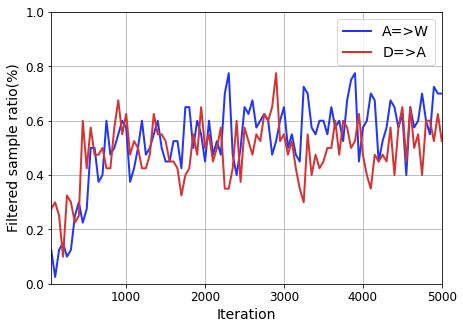} &
\includegraphics[width=0.6\linewidth]{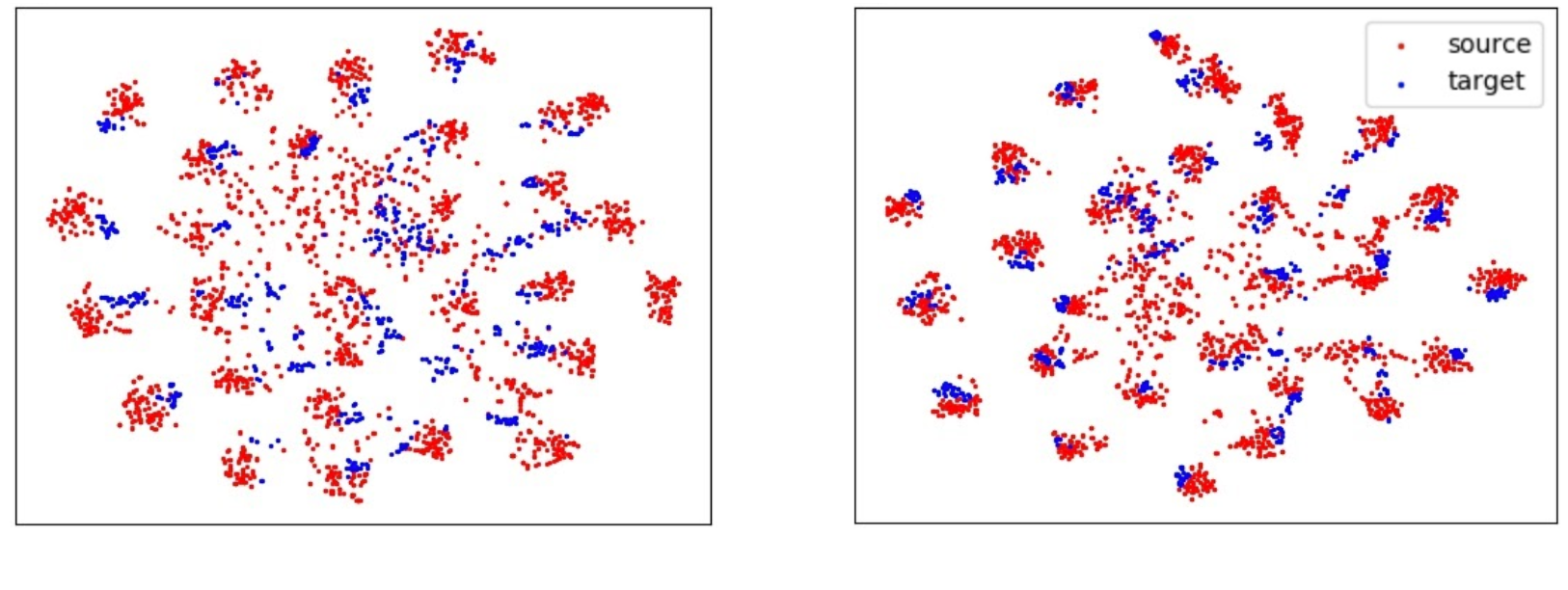}  
\\
{\hspace{4.3mm} (a) } & {(b) } \\
\end{tabular}
\vspace*{-0.05in}
\end{center}
\caption{
(a) Filtered sample ratio after confidence filtering with regard to training iteration.
(b) Feature space visualization of ResNet-50 (left) and the proposed SFDA (right).
}
\vspace*{-0.05in}
\label{fig:AW_experiments}
\end{figure}

\subsection{Empirical analysis}
\label{ssec:Analysis}

\textbf{Confidence-based Filtering:}
\label{sssec:conf_filter}
Our confidence-based filtering effectively addresses the uncertainty of the imperfect pseudo labels as shown in Table \ref{table:accuracy_officeic}, Table \ref{table:accuracy_officehome}, and Table \ref{table:accuracy_visda}.
To further validate our filtering scheme, we measure the percentage of valid training samples \ie $w=1$ in Eq. (\ref{eq: confidence_score}), on A $\Rightarrow$ W and D $\Rightarrow$ A as shown in Fig. \ref{fig:AW_experiments}(a).
We can see that a small portion of the target sample is used for training at the beginning, but the number of valid samples gradually increases as training progresses.

\textbf{Feature visualization:} 
\label{sssec:t_sne}
To visually examine the effectiveness of the proposed method, we compare t-SNE  embedding of the  features \cite{maaten2008visualizing} from ResNet-50 and our SFDA on {A} $\Rightarrow$ {W}  in Office-31. 
As shown in Fig. \ref{fig:AW_experiments}(b), we can observe that the source and target domains are better aligned by SFDA. This result clearly shows that it is possible to mitigate the discrepancy between two different domains without accessing source data.

\textbf{Statistics of reliable samples:}
\label{sssec:StatisticsofReliable}
We treat samples with the entropy values less than $0.2$ as ``reliable samples", which accounts for about 30\% of total samples as shown in Fig. \ref{fig:motivation}. 
Note that these statistics can change as our SFDA evolves progressively during the training phase.
Hence, we further analyze the changes in these statistics as training progresses.
From Fig. \ref{fig:WA_experiments}(a), we can observe the percentage of ``reliable samples" gradually increases and eventually exceeds 50\% of total samples. 
More importantly, the accuracy of pseudo-labels of reliable samples also increases.

\textbf{APM update period:}
\label{sssec:edge_type}
We update APM periodically to reflect the statistics of the target domain using our progressively evolving target model.
Figure \ref{fig:WA_experiments}(b) shows the performance of our method with different APM update periods.
The shorter the update period, the better the accuracy, but requires a higher amount of computation during training.
Conversely, increasing the update period (e.g. over $1000$) degrades performance as the target model cannot fully utilize its self-learning scheme.
Empirically, we set the update period to $100$ across all the datasets.

\textbf{Trade-off parameter $\alpha$:}
\label{sssec:tradeoff_alpha}
We analyze the trade-off parameter $\alpha$ between the source-regularization loss and the self-learning loss in Eq. (\ref{eq:overall_loss}) as shown in Fig. \ref{fig:WA_experiments}(c).
In the experiment, we investigate two strategies: static $\alpha$ and dynamic $\alpha$.
For static $\alpha$, we vary the value from $0$ to $1$. 
Using only source-regularization loss, \ie $\alpha = 0$, does not utilize the advantage of the updated target model, so performance is lower than in other settings.
On the other hand, relying solely on self-learning loss, \ie
$\alpha = 1$, may fall into local minima due to self-biasing.
In SFDA, we set dynamic $\alpha$ with a schedule, $\alpha = 2 (1+exp(-10 \cdot iter / max\_iter))^{-1}-0.5)$, so that the target model is updated gradually from the source model.
From this experiment, we can observe that dynamic $\alpha$  outperforms all settings of static $\alpha$, which demonstrates the effectiveness of our dynamic scheduling.

\vspace{-0.05in}
\section{Related work}
\vspace{-0.05in}

\textbf{Unsupervised domain adaptation (UDA):} 
Early UDA methods exploit distribution matching to alleviate the domain discrepancy between source and target domains.
Tzeng \etal \cite{tzeng2014deep} introduce a domain confusion loss using  maximum mean discrepancy (MMD)~\cite{tzeng2014deep,long2015learning}.
Subsequently, a number of variants based on distribution matching have been proposed, \eg JMMD~\cite{long2016unsupervised,long2016unsupervised}, CMD \cite{zellinger2017central}, MCD~\cite{saito2017maximum}. 
Meanwhile, there has been a growing interest in domain adversarial methods \cite{ganin2015unsupervised}.
In these approaches, the domain discriminator is trained to predict the domain label from features, and at the same time, the feature extractor tries to deceive the domain discriminator, resulting in domain-invariant features.
Due to its simple yet effective extensibility and outstanding results, domain adversarial training has been widely used across various fields \cite{bousmalis2016domain,tzeng2017adversarial,tsai2018learning,hong2020attention}.
Unfortunately, existing UDA methods require labeled source data, which can lead to privacy issues.
Instead of accessing the source data directly, we perform domain adaptation by leveraging a pre-trained source model.

\textbf{Fine-tuning:} One of the most practical paradigms in deep learning-based approaches is transfer learning. 
Especially, fine-tuning parameters of a pre-trained model on large-scale datasets~\cite{imagenet,Kinetics} to a specific target task has shown promising results in various areas \cite{karpathy2015deep,tzeng2017adversarial,radenovic2018fine}.
For instance, VGG~\cite{simonyan2014very} or ResNet~\cite{he2016deep} trained from ImageNet~\cite{imagenet} are extensively fine-tuned for   smaller dataset such as Pascal VOC~\cite{pascal-voc-2007,pascal-voc-2012}.
3D CNN models trained from Kinetics~\cite{Kinetics} are also used for UCF101~\cite{UCF101} and HMDB51~\cite{HMDB51} by default.
Furthermore, fine-tuning is widely applied in two different but highly-related tasks, \eg from image classification to object detection, and significantly outperforms a model trained from scratch.
Compared with the fine-tuning technique, our  SFDA protocol yields a crucial advantage of being free from labor-intensive labeling of the target domain. 

\textbf{Knowledge distillation (KD):} 
Our SFDA is related to KD~\cite{hinton2015distilling} in that there are two models in which one model depends on the other.
However, fundamentally, KD aims to transfer the knowledge of a large model (teacher) to a small model (student), so both models use the same dataset. 
In other words, there is generally no covariate shift \cite{sugiyama2007mixture} in KD.
Several studies~\cite{zagoruyko2017attention,srinivas2018distillation,heo2019distillation} have demonstrated that the teacher model pre-trained from the source data can transfer knowledge to the student model for the target domain.
Despite promising results even in the presence of domain discrepancy, these techniques all require the labeling of samples in the target domain.
In contrast, the proposed SFDA can transfer knowledge from a labeled source domain to an unlabeled target domain during training.

\vspace{-0.05in}
\section{Conclusion}
\vspace{-0.05in}
In this paper, we have proposed a novel paradigm shift for unsupervised domain adaption, called Source data-Free Domain Adaptation (SFDA) from a source pre-trained model.
Our main (target) model utilizes a pre-trained model from the source domain instead of using source data directly.
Specifically, two types of pseudo labels are used for training our target model.
\textit{Target-oriented pseudo labels} obtained from the adaptive prototype memory are used to train the target model in a self-learning manner while \textit{source-oriented pseudo labels} prevent the target model from a self-biasing problem.
Despite not directly accessing the source data, our model achieves higher performance than conventional models trained with source data.
From extensive empirical analysis for SFDA scenarios, we claim that SFDA is one of the effective ways to address data-privacy issues caused by labeled data with sensitive attributes.
{In this study, we address the closed-set domain adaptation scenario where the class set of source and target is identical.
But in real-world scenarios, the source and target class set might be different and it could affect our class prototype selection scheme.
As future work,  we plan to extend the proposed SFDA approach to more realistic openset and partial domain adaptation scenarios.}
Also, we will perform SFDA on datasets that contain sensitive labels, \eg  bio-metric information, and address the encountered issues.

\section*{Broader Impact}

This study suggests a paradigm shift by addressing the data-privacy issue, especially in unsupervised domain adaptation.
Based on our source data-free method, various stakeholders including enterprises and government organizations can be free of concern about privacy issues with their labeled source dataset.

Furthermore, the proposed data-free approach can contribute to creating a positive social impact, especially when the data is in large-scale.
Recently, since the size of data across various fields has been scaling up for model training, it is almost incapable for individual researchers,  startups, and small or medium enterprises to directly utilize such large scale of data during training.
For this reason, a new social trend of sharing pre-trained models, \eg NasNet, EfficientNet, and BERT, led by global enterprises with their huge amount of resources has been rising up.
These efforts can provide more opportunities for people to join AI community by participating AI-based research or related large-scale projects.
Our approach based on pre-trained models thus can create social impact by proposing new as well as beneficial solution, so that more people can participate in domain adaptation research specifically when the source data is large-scale and contains sensitive attributes.

On the other hand, our method potentially limits the degree of freedom when modifying or adding source data, \eg annotation, or data pre-processing, especially in the case of requiring/expecting more flexible domain adaptation.
In conclusion, we believe that SFDA proposes a new protocol in domain adaptation and even suggests a potential solution for increasing the need for safety in data privacy and data security.

\section*{Supplementary Material}

\begin{large}
\textbf{Appendix (A): Reimplementation}
\end{large}

In our experiments, we compare the proposed method with the previous UDA methods.
Importantly, early UDA methods report performance based on \textit{Caffe} while recent works are based on \textit{Pytorch} implementation.
Therefore, the performance of early works tends to be considerably lower than the recent works including ours.
For a fair comparison, we re-implement ResNet-50, DAN, DANN, and MSTN (DANN-based) using \textit{Pytorch}.

\textbf{Experimental setup}: 
We use a batch size of 32 across all the datasets.
Also, we set the maximum iteration step to 10,000 for Office-31, 20,000 for Office-Home.
We use SGD as the optimizer with a weight decay of 0.0005  and a momentum of 0.9.
The base learning rate is set to $10^{-3}$ and all fine-tuned layers in a pre-trained feature extractor are optimized with a learning rate of $10^{-4}$. 
We apply a learning rate schedule with $lr_p = lr_0(1+\alpha\cdot p)^{-\beta}$ where $lr_0$ is a base learning rate, $p$ is a relative step  that changes from 0 to 1  during training, $\alpha$ = 10, and $\beta$ = 0.75.

In Table \ref{table:reimplement_office31} and Table \ref{table:reimplement_officehome}, we compare the performance between reported accuracy (\textit{Caffe}) \cite{xu2019larger,long2018conditional} and our reimplementation (\textit{Pytorch}).
The results show that our re-implemented results achieve higher performance than the previously reported results, especially for ResNet baseline.
For a fair comparison, we reimplement early UDA methods using \textit{Pytorch} and compare them with the proposed method in our main paper.

\begin{table}[h!]
    \addtolength{\tabcolsep}{1.5pt}
    \centering
    \caption{Classification Accuracy (\%) on {Office-31}  ({ResNet-50})}
    \label{table:reimplement_office31}
    \resizebox{0.75\textwidth}{!}{%
    \begin{tabular}{lccccccccc}
        \toprule
        \multirow{2}{30pt}{\centering Method}\:\:\:\:\:\:\: &  \multicolumn{7}{c}{Office-31}  \\
        \cmidrule{2-8} 
        & \:A$\Rightarrow$W\: & \:D$\Rightarrow$W\: & \:W$\Rightarrow$D\: & \:A$\Rightarrow$D\: & \:D$\Rightarrow$A\: & \:W$\Rightarrow$A\: & \:Avg\:   \\
        \midrule
        ResNet  (reported)   &  68.4 $\pm$ 0.2& 96.7 $\pm$ 0.1& 99.3 $\pm$ 0.1& 68.9 $\pm$ 0.2& 62.5 $\pm$ 0.3& 60.7 $\pm$ 0.3& 76.1  \\
        DAN  (reported)       &80.5 $\pm$ 0.4 & 97.1 $\pm$ 0.2 & 99.6 $\pm$ 0.1 & 78.6 $\pm$ 0.2 & 63.6 $\pm$ 0.3 & 62.8 $\pm$ 0.2 & 80.4   \\
        DANN (reported) & 82.0 $\pm$ 0.4 & 96.9 $\pm$ 0.2 & 99.1 $\pm$ 0.1 & 79.7 $\pm$ 0.4 & 68.2 $\pm$ 0.4 & 67.4 $\pm$ 0.5 & 82.2   \\
        \midrule
        ResNet   & 79.8  $\pm$  0.6 & 98.3  $\pm$  0.3  & 99.9  $\pm$  0.1 & 83.8  $\pm$  0.5 & 66.1  $\pm$  0.3 & 65.1  $\pm$  0.2 & 82.2  \\
        DAN       & 82.6  $\pm$  0.6 & 98.5  $\pm$  0.1  & 100.0  $\pm$  .0 & 83.6  $\pm$  0.5 & 67.3  $\pm$  0.8 & 66.3  $\pm$  0.4 & 83.1   \\
        DANN & 84.8  $\pm$  0.9 & 98.2  $\pm$  0.3  & 99.9  $\pm$  0.1 & 86.3  $\pm$  0.8 & 68.9  $\pm$  0.4 & 66.8  $\pm$  0.6 & 84.1   \\
        \bottomrule
    \end{tabular}%
    }
 \vspace*{-0.1in}
\end{table}

\begin{table}[h!]
    \addtolength{\tabcolsep}{1.5pt}
    \centering
    \caption{Classification Accuracy (\%)  on {Office-Home}  ({ResNet-50})}
    \label{table:reimplement_officehome}
    \resizebox{1\textwidth}{!}{%
    \begin{tabular}{lccccccccccccc}
        \toprule
        \multirow{2}{30pt}{\centering Method}\:\:\:\:\:\:\: & \multicolumn{13}{c}{Office-Home} \\
        \cmidrule{2-14}
        & {Ar}$\Rightarrow${Cl} & {Ar}$\Rightarrow${Pr} & {Ar}$\Rightarrow${Rw} & {Cl}$\Rightarrow${Ar} & {Cl}$\Rightarrow${Pr} & {Cl}$\Rightarrow${Rw} & {Pr}$\Rightarrow${Ar} & {Pr}$\Rightarrow${Cl} & {Pr}$\Rightarrow${Rw} & {Rw}$\Rightarrow${Ar} & {Rw}$\Rightarrow${Cl} & {Rw}$\Rightarrow${Pr} & \:\:Avg\:\: \\
        \midrule
        ResNet (reported)          & 34.9 &50.0&58.0&37.4&41.9&46.2&38.5&31.2&60.4&53.9&41.2&59.9&   46.1 \\
        DAN (reported)       & 43.6 &57.0 &67.9 &45.8 &56.5 &60.4 &44.0 &43.6 &67.7 &63.1 &51.5 &74.3 &   56.3 \\
        DANN (reported) & 45.6& 59.3& 70.1& 47.0& 58.5& 60.9& 46.1& 43.7& 68.5& 63.2& 51.8& 76.8 &   57.6 \\
        \midrule
        ResNet           & 42.5 & 65.8 & 74.1 & 57.4 & 63.7 & 67.7 & 55.7 & 39.1 & 72.9 & 66.1 & 46.3 & 76.9 & 60.7 \\
        DAN       & 45.4 & 65.8 & 73.9 & 56.9 & 61.4 & 65.9 & 56.1 & 43.0 & 73.2 & 67.5 & 50.4 & 78.5 & 61.5 \\
        DANN & 45.7 & 66.6 & 73.4 & 58.0 & 64.9 & 68.3 & 55.9 & 42.6 & 74.1 & 66.5 & 49.7 & 77.5 & 62.0 \\
        \bottomrule
    \end{tabular}%
    }
 \vspace*{-0.1in}
\end{table}

\vspace{6mm}
\begin{large}
\textbf{Appendix (B): Additional ablation on VisDA-C}
\end{large}

We perform the ablation analysis regarding a trade-off parameter $\alpha$ on VisDA-C.
A detailed description of the experimental protocol is described in Section 3.3 in our main paper.
As  reported in Table \ref{table:alpha_visda}, we investigate two strategies: static $\alpha = [0.0, 0.2, 0.4, 0.6, 0.8, 1.0]$ and dynamic $\alpha = 2 (1+exp(-10 \cdot iter / max\_iter))^{-1}-0.5)$.
From the table, we can observe similar results as in Office-31.
Using only source-regularization loss ($\alpha = 0$) or self-learning loss ($\alpha = 1$)  degrades performance since they rely on one-sided information (source or target).
Although using both source and target information shows promising results in the static setting, dynamic $\alpha$  outperforms all the settings of static $\alpha$.

\begin{table}[h!]
    \addtolength{\tabcolsep}{1.5pt}
    \centering
    \caption{Classification Accuracy (\%)  on {Visda-C}  ({ResNet-101})}
    \label{table:alpha_visda}
    \resizebox{1\textwidth}{!}{%
    \begin{tabular}{lccccccccccccc}
        \toprule
        \multirow{2}{30pt}{\centering Method}\:\:\:\:\:\:\: & \multicolumn{13}{c}{Vidsa-C} \\
        \cmidrule{2-14}
        & aero & bicycle & bus & car & horse  & knife & motor & person & plant & skate & train & train & \:\:Avg\:\: \\
        \midrule
        {{SFDA ($\alpha = 0.0$)}}            & 87.1 & 66.7 & 73.4 & 56.2 & 91.1 & 43.9 & 83.3 & 57.8 & 76.8 & 53.1 & 84.7 & 43.0  & 68.1 \\
        {{SFDA ($\alpha = 0.2$)}}           & 87.9&	70.0&	80.5&	63.8	&92.2&	66.1&	87.9&	65.0&	78.3&	56.9&	78.0	&46.6&	72.8 \\
        {{SFDA ($\alpha = 0.4$)}}            & 88.5	&75.2&	81.9&	62.7	&91.5&	82.3&	88.0&	68.8&	81.9&	62.3	&77.4&	44.5&	75.4 \\
        {{SFDA ($\alpha = 0.6$)}}            & 86.5	&74.8&	81.2&	61.0	&90.4&	90.8&	84.7&	77.6&	75.9&	52.7&	78.2&	47.7&	75.1\\
        {{SFDA ($\alpha = 0.8$)}}            & 85.5	& 77.0	&84.4	&62.9	& 91.8	& 89.6	& 81.8	&76.4	&78.4	&54.6	&76.5	&32.3	&74.3 \\
        {{SFDA ($\alpha = 1.0$)}}            &86.8	&81.2	&86.5	&60.6	&90.7	&91.3	&82.3	&69.5	&81.5	&51.6	&73.0	&37.6	&74.4 \\
        \midrule
        {{SFDA} (dynamic $\alpha$
        )}            & 86.9 & 81.7 & 84.6 & 63.9 & 93.1 & 91.4 & 86.6 & 71.9 & 84.5 & 58.2 & 74.5 & 42.7  & 76.7 \\
        \bottomrule
    \end{tabular}%
    }
\end{table}

\vspace{6mm}
\begin{large}
\textbf{Appendix (C): Confidence-base filtering}
\end{large}

In this section, we introduce Hausdorff distance (Eq. 3) in Confidence-base filtering in more detail.
The objective of confidence-base filtering is to estimate the confidence of a pseudo label by considering the corner cases between two sets.
Here, we compute the relationship of three sets as following:

 $\bullet$ $Q$: a singleton set containing embedded feature of sample $x_t$ , \ie $\left\{ f_t \right\}$.
 
  $\bullet$  $M_{t1}$: a set containing multiple prototypes of the most similar class (defined by $s_c (x_t)$ in Eq. 2).
 
 $\bullet$ $M_{t2}$: a set containing multiple prototypes of the second similar class.

The Hausdorff distance can be formulated as follows:
\begin{equation}
    \begin{split}
       d_{H}(Q, M_{t1}) & = \max \left\{ 
    \adjustlimits\sup_{q\in Q} \inf_{p\in M_{t1}} {d}(q,p),   
    \adjustlimits\sup_{p\in M_{t1}} \inf_{q\in Q} {d}(q,p)
    \right\} \\
     & = \max \left\{ 
    \adjustlimits\max_{q\in Q} \min_{p\in M_{t1}} {d}(q,p),   
    \adjustlimits\max_{p\in M_{t1}} \min_{q\in Q} {d}(q,p)
    \right\} \\
     & = \max \left\{ 
    \adjustlimits \min_{p\in M_{t1}} {d}(f_t ,p),   
    \adjustlimits\max_{p\in M_{t1}}  {d}(f_t ,p)
    \right\} 
    = \max_{ p\in M_{t1}} {d}(f_t ,p).
    \end{split}
\end{equation}

In the second line, since $Q$, $M_{t1}$, and $M_{t2}$ are closed and discretized sets, 
we can convert the \textit{suprema} and \textit{infima} to \textit{maxima} and \textit{minima},
respectively. 
In the third line, since $Q$ is a singleton set, \ie $f_t$ is the only element in the set, we can simplify $max$ and $min$   by inserting $f_t$ to $q$.

In the similar way, for the second most similar class, we slightly modify Hausdorff distance to select the opposite corner case as follows:
\begin{equation}
    \begin{split}
       d_{H}^* (Q, M_{t2})& = \min \left\{ 
    \adjustlimits\sup_{q\in Q} \inf_{p\in M_{t2}} {d}(q,p),   
    \adjustlimits\sup_{p\in M_{t2}} \inf_{q\in Q} {d}(q,p)
    \right\} \\
    & = \min \left\{ 
    \adjustlimits\max_{q\in Q} \min_{p\in M_{t2}} {d}(q,p),   
    \adjustlimits\max_{p\in M_{t2}} \min_{q\in Q} {d}(q,p)
    \right\}  \\
    & = \min \left\{ 
    \adjustlimits \min_{p\in M_{t2}} {d}(f_t ,p),   
    \adjustlimits\max_{p\in M_{t2}}  {d}(f_t ,p)
    \right\} 
    = \min_{p\in M_{t2}} {d}(f_t,p).
    \end{split}
\end{equation}

\bibliographystyle{splncs}
\bibliography{ref}
\end{document}